\documentclass[sigconf,nonacm]{acmart}
\AtBeginDocument{%
  }

\setcopyright{acmlicensed}
\copyrightyear{2026}
\acmYear{2026}
\acmConference[CHI '27]{CHI '27}{May 10--14, 2027}{Pittsburgh, PA}
\acmISBN{978-1-4503-XXXX-X/2018/06}

\usepackage{quoting}
\usepackage{xcolor}
\quotingsetup{leftmargin=1em, rightmargin=1em, font=small}
\definecolor{sarah}{RGB}{34, 139, 34}
\definecolor{CEK}{RGB}{41, 98, 204}

\begin{document}

\title[The House with a Million Windows]{The House with a Million Windows:\\Interactive Fiction for Narrative Restorying}


\author{Cody Kommers}
\authornote{Both authors contributed equally to this research.}
\email{ckommers@turing.ac.uk}
\orcid{https://orcid.org/0009-0007-8985-0085}
\affiliation{%
  \institution{The Alan Turing Institute}
  \city{London}
  \country{United Kingdom}
}

\author{Sarah Immel}
\authornotemark[1]
\email{s.immel@ed.ac.uk}
\orcid{https://orcid.org/0009-0003-3778-4988}
\affiliation{%
  \department{CDT in Designing Responsible NLP}
  \institution{University of Edinburgh}
  \city{Edinburgh}
  \country{United Kingdom}}

\author{Drew Hemment}
\orcid{https://orcid.org/0000-0002-0068-5500}
\affiliation{%
  \institution{The Alan Turing Institute}
  \city{London}
  \country{United Kingdom}
}
\affiliation{%
  \institution{University of Edinburgh}
  \city{Edinburgh}
  \country{United Kingdom}
}
\author{Mina Lee}
\orcid{https://orcid.org/0000-0002-0428-4720}
\affiliation{%
  \institution{University of Chicago}
  \city{Chicago}
  \country{United States}}

\renewcommand{\shortauthors}{Kommers, Immel et al.}

\begin{abstract}


AI-assisted writing can flatten meaning in human storytelling, enabling the production of homogeneous outputs without the intentional effort and sense-making writing entails. To address this challenge, we present The House with a Million Windows (HWAMW), an LLM-based interactive fiction system designed to help users explore both the breadth and depth of potential meanings within their personal stories---drawing on a psychological paradigm called the restorying intervention. In HWAMW, users play through a text-based narrative in which they tell a story, then encounter a set of LLM-generated ``windows'' reframing it according to different literary styles. Empirical evidence shows that HWAMW increases users' sense of narrative identity, while an expert review explores how this effect is achieved. Our findings suggest that HWAMW facilitates restorying and offers a valuable paradigm for AI-assisted writing, wherein LLMs do not tell our stories but rather help us see greater potential in the stories we tell. 

\end{abstract}


\ccsdesc[500]{Human-centered computing~Natural language interfaces}
\ccsdesc[300]{Human-centered computing~User studies}
\ccsdesc[300]{Human-centered computing~Empirical studies in HCI}

\keywords{interactive fiction, co-creativity, narrative, restorying, meaning, meaning-making, metaphor, interpretive technologies}

\begin{teaserfigure}
 \includegraphics[width=\textwidth]{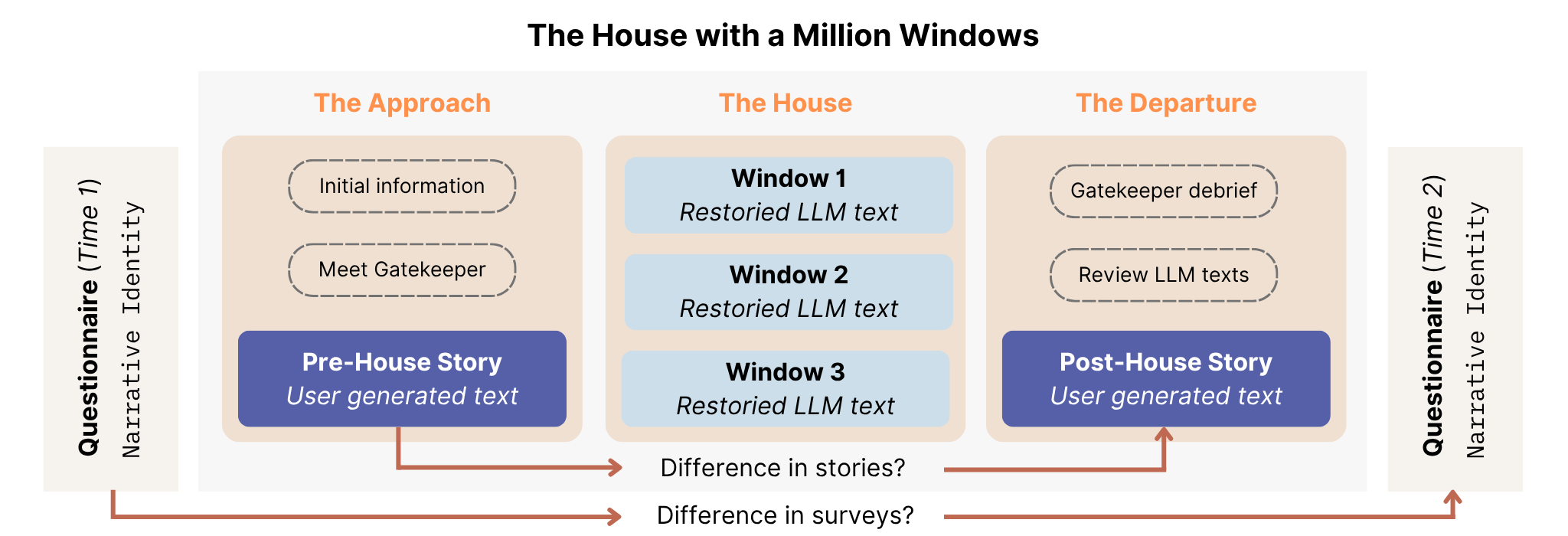}
  \caption{\textit{The House with a Million Windows prototype}. Users experience a metafictional narrative which guides them through reflecting on a personal story or life event. \textit{The Approach}. Within the narrative, users provide information about themselves and meet a character called the Gatekeeper, to whom they tell an initial version of their story. \textit{The House}. Users then explore several ``windows'' with LLM-generated texts representing possible variations of their story based on pre-defined literary styles and techniques. \textit{The Departure}. They end by debriefing with the Gatekeeper, during which they are allowed to review the LLM variations while telling a new version of their story. For empirical validation, users filled out a questionnaire on narrative identity before and after experiencing the prototype, allowing us to compare changes in users' stories and survey responses.}
  \Description{A schematic with a main bank for the House with a Million Windows. It includes three sub-banks for The Approach, The House, and The Departure. The Approach shows where the user writes their first story. The House shows where the user encounters three ``windows'' with restoried LLM-generated texts. The Departure shows where the user writes their second story. Outside the main bank, there are sub-banks for questionnaires, both before and after the House with a Million Windows bank. Arrows show that the questionnaires can compared differences, as can the Pre- and Post-House user-generated stories.} 
  \label{fig:teaser}
\end{teaserfigure}


\maketitle


\newpage

\begin{quote}
    \begin{center}
        
    ``The house of fiction has in short not one window, but a million'' -- \textit{Henry James} \cite{james1908preface}

    \end{center}

\end{quote}

\section{Introduction}

Storytelling is a core part of the human experience: the narratives we tell play a central role in how we make sense of time \cite{ricoeur1980narrative}, ourselves \cite{bruner1987life}, and the world around us \cite{bruner1991narrative}. Generative AI has the potential to fundamentally alter or even displace this process, as large language models (LLMs) make the generation of narrative text possible without the effort and intention of human storytelling. Systems must therefore be explicitly designed with the intention to think \textit{with} humans rather than for them \cite{collins2024building}. While many of the implications of this trend have been explored, it is still unclear how this will affect human narrative practices and individuals' sense of identity and meaning. If we outsource the process of writing to AI, do we also delegate the construction of identity?

We respond to this question by addressing a specific risk associated with AI-generated writing---the flattening of meaning \cite{kommers2026computational} in both the writing process and the narratives that are written. Outputs composed or edited with AI assistance tend to be more culturally or stylistically homogeneous \cite{heuser2025cultural, xie2026artificial, veselovsky2026localized}. Eliminating the cognitively effortful process of writing removes an important opportunity to ``think through'' what one is writing about, particularly when it comes to discovering or constructing new layers of meaning from ones own experiences \cite{kommers2025sense, zohar2026against, lee2025impact}. But this risk reflects the initial circumstances of generative AI's development and deployment, rather than an intrinsic fact of the technology itself \cite{kommers2025meaning, collins2024building, kommers2026slop}. Integrating these technologies in ways which enrich writing processes is an established design problem \cite{booten2024build}, one we consider specifically with respect to the formation and development of personal narratives. We consider that LLMs, though often employed in ways which automate and displace human stories, might instead provide new opportunities for people to slow down, tell, and retell their stories, expanding not only the space of what we write but also the personal and cultural meanings we create in the process \cite{vara2026searches}.


\begin{figure}
  \centering
  \includegraphics[width=\columnwidth]{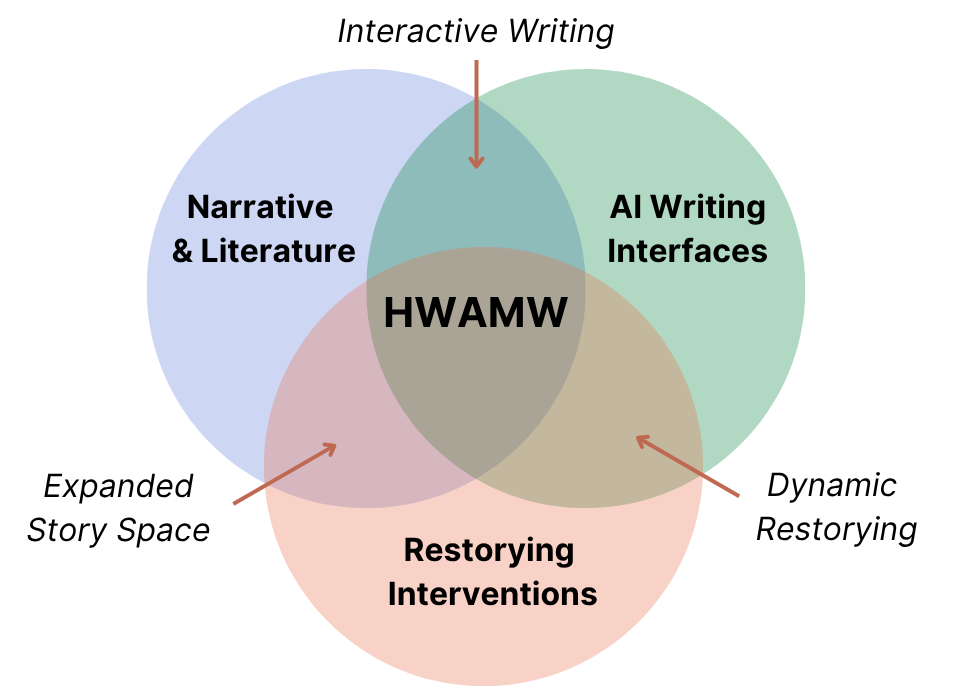}
  \caption{The House with a Million Windows interface combines approaches from narrative and literature, generative AI writing interfaces, and restorying interventions. Each of these overlaps has been explored in existing work. The novel aim of HWAMW is to offer a unique combination drawing on all three of these perspectives.}
  \Description{Three overlapping circles in a Venn diagram. The circles are labelled: Narrative and literature, Generative AI writing Interfaces, and Restorying intervention. Each of their intersections has a label: interactive writing, dynamic restorying, and expanded story space. In the middle is HWAMW.}
  \label{fig:venn}
\end{figure}


In this paper, we introduce The House with a Million Windows (HWAMW), a research prototype through which we investigate how LLM-generated variations can support narrative restorying while preserving users' authorship and agency. In the HWAMW prototype, users experience a metafictional narrative which guides them through the process of restorying, eliciting their initial story and prompting them to reflect on variations and possible retellings before asking them to tell it again, consolidating new perspectives as they consider how their story and its meaning have changed. This interface is based in psychological and literary practices of ``restorying'' \cite{rogers2023seeing}---recasting the same series of events with different narrative choices to explore new perspectives, interpretations, and meanings. We evaluate the HWAMW prototype through two complementary studies: (1) an empirical validation with general users quantitatively examining changes in narrative identity and written narratives before and after interacting with the system \ref{empiricalValidation}; and (2) an expert review qualitatively examining the initial implementation of our approach to elicit third-party perspectives on design considerations or future directions. The prototype contributes to work in the emerging area of Interpretive Technologies \cite{hemment2025doing, kommers2026computational} and sits at a novel intersection of three strands of research: narrative theories and practices, particularly interactive fiction \cite{murray1997hamlet}; AI-driven interfaces for interactive writing \cite{lee2024design}); and restorying interventions from experimental psychology \cite{rogers2023seeing} (Fig.~\ref{fig:venn}). 

Restorying can be a psychologically beneficial process, potentially helping users to deconstruct counter-productive narratives and find a greater sense of meaning or control in their stories \cite{singer2004narrative, mcadams2011narrative, de2024identity}. However, HWAMW is designed as a reflective writing interface rather as a clinical therapeutic aid. The possibilities of narrative-driven restorying are both more limited and more general than what takes place in a therapeutic setting: more limited, in that they do not come with the clinical implications associated with therapy, including expertise and oversight of a licensed practitioner; more general, in that narrative restorying can open up a vast range of individual or cultural meanings, only a subset of which are likely to be targeted in therapy, such as those available in works of art and design \cite{immel2026stepping, hemment2024experiential, hemment2025sensing}. Avoiding the appearance of a ``therapy bot'' is thus a key design constraint in our work.

Our goal is to emulate something closer to a creative writing workshop or literary adaptation. HWAMW is designed to help users who have a story to tell become more aware of the range of narrative techniques and possible tellings available to them. Without assuming any narrative expertise or training, it allows them to explore narrative structures, devices, and strategies they might apply to a personal story by re-encountering it through procedurally generated text in a way which is engaging, creative, and psychologically constructive. More concretely, our aims in designing this system are threefold: (a) to leverage LLM-generated text to support users to engage in exploratory restorying, while preserving their authorship and agency in writing their own stories, (b) to draw on literary fiction to expand the range of narrative techniques and approaches which might be considered as valuable restorying interventions, and (c) to explore the psychological impacts of narrative restorying, while being careful to avoid any implicit suggestions of clinical effects the system is not designed to fulfil.

This paper makes three contributions:
\begin{enumerate}
\item We introduce metafictional restorying as an approach to AI-assisted personal writing in which LLM-generated retellings are framed as ``windows'' onto a user's own story, offering possibles views to look through rather than drafts to accept, as instantiated in the HWAMW prototype (Section~\ref{sec:HWAMW}).
\item We provide initial quantitative evidence that a single session with HWAMW shifts how users narrate a personal experience, strengthening self-reported narrative identity and measurably changing the structure of their retold stories (Section~\ref{empiricalValidation}).
\item We offer a qualitative account of the opportunities and tensions of metafictional restorying through an expert review, characterizing how HWAMW’s narrative framing, literary interventions, and imperfect LLM outputs can support or disrupt reflection, voice, and meaning-making, revealing both the promise and limitations of this approach (Section~\ref{expertReview}).

\end{enumerate}

We connect this work to key theories and related work in literary studies, narrative psychology, and human-computer interaction (Section \ref{relatedWork}), then introduce the design of our system, highlighting the ways it expands psychological restorying interventions (Section \ref{restoryingRedesigned}) and reframes LLM contributions to personal writing (Section \ref{AIReframed}). We detail HWAMW's implementation and user experience (Sections \ref{architecture}--\ref{ux}) and empirically validate its narrative and restorying effects with crowdsourced users from a general audience (N=40). While our empirical evidence shows that participants' narratives and narrative identities shifted over their experience with the interface (Section \ref{empiricalValidation}), we sought to better qualify and improve that experience through a focused expert user review (N=10; Section~\ref{expertReview}). Our qualitative analysis suggests that HWAMW's dynamic restorying and metafictional approach are valuable for meaning-making in personal writing and that the chosen narrative interventions and LLM-generated ``window'' texts are well motivated but could be expanded in future work. We discuss the value of HWAMW's general approach to AI-assisted writing as a means for making more meaningful narratives with and about AI (Section \ref{discussion}) and highlight opportunities to build and expand on this work (Section \ref{sec:limits-and-future}).

\section{Related Work} \label{relatedWork}

\subsection{Restorying}

When trying to make sense of a complex reality, people often seek to organise their experience into narratives \cite{bruner1991narrative, bruner1987life, kommers2025sense}. Psychologists have long been interested in codifying the stories people tell stories about themselves and what psychological effects these stories might have on their perception of meaning in life or sense of identity \cite{bruner1990acts, allport1965letters, bruner1994remembered, mcadams1995what, mcadams2011narrative, mcadams2001psychology}. This psychological impact of framing and re-framing stories has perhaps most famously been posed as a kind of search problem---traversing possible narrative re-framings in an effort to maximise one's sense of meaning \cite{frankl1959mans}. 

However, there is often a disconnect between this empirical work on the psychology of stories and the richness of storytelling as it is understood in disciplines like literary studies and creative writing. If literature is like a house with a million windows \cite{james1908preface}, then not only are each of these different windows characterised by myriad subtle differences (e.g., authorial style, structure, themes, genre conventions), they also depend on the interpretation of the person peering through them. Most psychological research makes limited contact with this richer sense of narrative informed by literary fiction.

\subsubsection{In literature and creative writing}

Restorying, considered more broadly as the reshaping of dominant or accepted narratives, has a longer history in literature and the arts, where ancient stories and myths have frequently been adapted, interpreted, and altered to challenge traditions, reach new audiences, or speak to present-day experiences and concerns \cite{hutcheon2006theory}. In recent work, the term `restorying' has been applied to a broad range of re-tellings: \textit{recycling} existing narratives to create something new, significantly \textit{recontextualising} a narrative while retaining much of its content or structure, and \textit{reinterpreting} stories with an emphasis or obligation toward fidelity \cite{scott2020restorying}. Of these, \textit{reinterpretation} maps most closely to existing work in psychological restorying: the events, sequence, and lived experience underlying personal narratives remains the same, while discourse, tone, and the style of telling may shift to communicate the story differently.

However, each form of re-telling has the potential to shape the meaning of the original text: all texts are interpreted in light of other texts (a property commonly referred to as \textit{intertextuality}) \cite{kristeva1967word}. Adaptations and re-tellings create especially strong connections. They both represent versions of a story that has been retold---and so is no longer understood in isolation, but carries additional layers of interwoven meaning \cite{hutcheon2006theory}. Personal restorying then becomes something much broader: personal narratives are already shaped by ancient stories and myths, fictions and fantasies; making these connections stronger through adaptation and retelling facilitates not just personal meaning but cultural, artistic, and historic meaning-making as well. 

Sometimes, this restorying and intertextuality is itself explored in literature, as in classic works of metafiction such as ``The Library of Babel'' (a story about a library containing every possible enumeration of a fixed-length text) and ``The Garden of Forking Paths'' (a story concerning a labyrinthine novel which explores branching possible futures) by Jorge Luis Borges \cite{borges1999collected}. This traditional has been carried into more recent metafictional works, like \textit{The Midnight Library} by Matt Haig, in which the protagonist enters a library where each book she pulls from the shelf offers a view into different versions of her life. As these works demonstrate, restorying is not only something that happens on top of stories, as an epiphenomenal experience of the reader; it can happen within them as well.

We draw on this vein of literary and media theory to motivate our expanded view of potential restorying interventions, moving beyond \textit{reinterpretations} where fidelity is the main or only goal to \textit{recycle} and \textit{recontextualise} as well, and setting this restorying work in a metafictional frame, where stories and the relationships between them become easier to explore.

\subsubsection{Psychological effects of restorying}

In experimental psychology, the restorying intervention is a paradigm for encouraging people to draw on narrative structures and techniques in narratives about their life or experience \cite{rogers2023seeing, mcadams2011narrative, koehler2025make}. In a typical restorying experiment, participants are asked to give a baseline (Time 1) narrative, after which they are provided with a series of prompts to emphasise how their story might align with a particular structure or theme. They are then asked to offer a restoried (Time 2) version of their narrative. After each story, they can fill out a questionnaire designed to measure a particular psychological effect; the intersection succeeds if a measurable differences can be seen between the Time 1 and Time 2 narratives or questionnaires.

The crucial feature of the restorying intervention is that it does not alter the events experienced by participants---but rather the story they tell about them. It demonstrates how the psychology of meaning depends not on some kind of intrinsic meaningful value of the events themselves but a specific interpretation of them \cite{kommers2025sense}. This paradigm offers a vehicle for studying concrete psychological effects of a range of narratological structures and approaches. Specifically, previous work show that a restorying intervention based on the ``hero's journey'' can increase participants' perceived meaning in life, as measured by a self-report scale \cite{rogers2023seeing}. 

There are two significant limitations to this previous work. The first is that it focuses on a limited range of story structures---specifically, the hero's journey \cite{campbell1949hero}. While widely known and cited, this is only one story structure among many. The second limitation is that restorying as a behavioural experiment relies on static prompts. While useful for codifying specific psychological effects, it is a limitation for tapping into the full power and flexibility of the narrative form. We build on the experimental approach developed by Rogers et al. \cite{rogers2023seeing} to prioritise a wider range of storytelling techniques and a more dynamic approach to restorying.

\subsection{Large Language Models and Narrative Writing}

LLMs are increasingly used to produce or assist in the production of narratives, from short stories and scripts to fanfiction, roleplay, and erotica \cite{gupta2026ai, ghajargar2022redhead, weber2024wraiter}. AI-generated narratives can read as highly convincing natural texts: as early as 2020, Elkins and Chun considered that selective GPT-3 outputs could pass as believable student work \cite{elkins2020can}, and models have grown more convincing ever since. However, LLM-generated narratives have been shown to be distinct from human writing---not only in rhetorical style \cite{reinhart2025llms} and stylistic diversity \cite{o2025stylometric}, but also in narrative choices such as neat, linear plotlines and explicitly moralising themes \cite{russell2026storyscope}. The threat of these technologies to replace human writing, decrease literacy, and homogenise creative outputs is a major concern for writers and researchers alike \cite{collett2025impact, baron2023wrote}.

\subsubsection{AI Writing Assistants}

Many HCI researchers have therefore turned their attention to uses of LLMs which may support, rather than replace, human writing. Lee et al. \cite{lee2024design} present a design space which covers 115 papers focused on user interactions with already-existing ``intelligent and interactive writing assistants,'' and many more have been developed since. Such systems may assist at various points in the writing process, from ideation to drafting or revision; many popular writing assistants, such as Sudowrite, integrate AI assistance at several distinct stages.

Writers working with AI writing assistants have found it helpful for accessibility \cite{loewen2024imagination}, productivity \cite{lee2022coauthor}, and creative expression \cite{wan2024felt} in ways not easily captured by commonly-suggested roles like ``tool'' or ``collaborator'' \cite{guo2025pen}. However, many writers consider the use of writing assistants ethically fraught, and their reception of the technology is highly individualised, affected by users' perceptions of the system and their values, especially creativity and authenticity \cite{gero2023social}.

\subsubsection{LLM Interfaces for Personal Writing}

AI-assisted writing---already a personal, values-driven, and highly contested process---becomes even more sensitive in the context of personal writing. Many new systems for AI-assisted reflective writing have been proposed in recent years, from assistants which generate journaling prompts, such as MindFeed \cite{10.1145/3786995.3787053}, to those directly assist in the initial drafting, like DiaryMate \cite{10.1145/3613904.3642693}, and, finally, those aiding with editing and reflection. Of these, ExploreSelf \cite{10.1145/3706598.3713883}, a journaling system which requires the user to write an entry before then presenting AI-generated reflection questions and themes, falls closest to our work here. However, each of these systems aims to rework past experiences toward positive themes and conclusions, focusing on reshaping the user's state of mind and not necessarily their story. While restorying based on literary fiction can have psychological benefits, it is not explicitly teleological in the same way. For example, literary fiction often dwells on the issues, complications, and negatively valenced aspects of a situation, rather than pushing towards a positive resolution.

We seek to challenge and extend this vein of work by exploring not just reflective journaling but personal writing and creative non-fiction. The key difference is that personal and creative writing has an indirect effect on the writer's mental states---as opposed to using writing instrumentally to achieve a specific desired mental state. It achieves this indirect effect (in many, though by no means all, cases) by connecting the writer to a wide range of narrative, artistic, and cultural meanings. Our work positions personal narratives as valuable in and of themselves, not only as a proxy for a user's state of mind.


\section{The House with a Million Windows}
\label{sec:HWAMW}

In The House with a Million Windows (HWAMW), users restory their personal narratives as part of a larger interactive experience drawing on techniques from works of metafiction. This container narrative prompts them to consider and tell a life story that feels unresolved or still troubles them, then presents them with fictionalised LLM-generated variations of their own story which are reframed based on established narrative techniques or well-known styles. After interacting with these restoried narratives, users are prompted to reflect on what they have seen and write their story once more, using their own words.

In this section, we discuss the design considerations, influences, and decisions involved in creating an engaging and interactive application for narrative restorying (Section \ref{restoryingRedesigned}) which uses LLM-generated text to support personal meaning-making and deepen user agency over their stories (Section \ref{AIReframed}). We then describe the technical implementation of the system (Section \ref{architecture}) and summarise the narrative as a user would encounter it (Section \ref{ux}).

\subsection{Restorying Interventions, Redesigned} \label{restoryingRedesigned}
Research through design offers a promising opportunity to not only study restorying but also to facilitate it in an engaging and accessible way. We saw several opportunities to design richer and more engaging restorying interventions:

\begin{enumerate}
    \item \emph{Metafiction as a context for restorying.} 
    An interface that simply suggests ways a user can restory their narratives comes across as a form of therapy. A key goal for our system was to design an interface for restorying that takes care to avoid putting the user in the mindset of a clinical setting. To accomplish this, we use metafictional narrative devices to differentiate this interface from AI-assisted therapy.
    
    \item \emph{Expanded story space.} 
    The space of possible restorying frames is unbounded; the interface should flexibly accommodate an arbitrary large number of them. While we only implement five possible windows in our prototype, it has the potential to incorporate an arbitrarily wide range of both stylistic and structural literary interventions.
    
    \item \emph{Dynamic, rather than static, restorying.}
    In creative writing practice, it is common for writers to learn new techniques by both reading and writing in that mode or style, often in an interactive community setting where they can receive personalised guidance and feedback. This experience is far removed from existing psychological restorying paradigms, which are designed for experimental precision---the ability to isolate a single variable and study its effects. We seek to retain useful elements of a rigorous experimental paradigm, while prioritising the playful, flexible, and engaging aspects of creative writing.
    
    \item \emph{In-story measures of meaning.}
    Our system features multiple ways of measuring how stories change over the course of restorying---and what psychological effects this might entail. In the context of a study, participants can fill out targeted questionnaires before and after engaging with the system (as in the psychological paradigm). But we also designed an in-game analog of the quantitative measure of meaningfulness---with a character called the Gatekeeper asking users for an explicit rating of how meaningful they consider their baseline and final stories to be.
    
    \item \emph{Multiple and iterative retellings.} 
    Instead of unsettling users' pre-narratives by prompting them to rework them according to a single fixed pattern, we encourage users to explore a broader selection of possible tellings and to chose which narrative interventions, if any, they incorporate into their retelling.
\end{enumerate}

\subsection{Reframing AI assistance in personal writing} \label{AIReframed}
Generative AI greatly expands the potential to achieve the above aims: prompting an LLM allows us to present variations of a user's story far more rapidly than rewriting it by hand, and far more flexibly than previous techniques for procedural text generation would have afforded. However, introducing this level of interactivity---and LLM use in particular---into the restorying of narratives of a potentially sensitive or vulnerable nature required us to take extra care in designing for user agency, framing AI-generated stories in a way that could maintain users' sense of authorship and authenticity throughout the restorying process.

\subsubsection{Metafictional Framing} \label{metafiction}

Games and interactive narratives make players the co-authors of their particular encounter within a designed story structure, giving them a stronger sense of agency \cite{murray1997hamlet}. In the restorying of personal narratives, this is especially desirable: users must feel some level of control over their narratives in order to meaningfully alter them. In order to support users' agency while also scaffolding them through the difficult and potentially vulnerable process of writing their own stories, we chose to frame the restorying experience within a larger container story or \textit{meta-narrative}. 

HWAMW is itself a narrative the user enters into as a character. Within this fictional setting, they tell their story to a mysterious Gatekeeper, then walk into a house where every window represents a possible telling of their story they might explore. By navigating the House on their own and completing key moments of the metanarrative with their own open-ended or multiple choice selections, HWAMW users become active participants in the restorying process. The fictional setting empowers them to coauthor the frame story and also creates a comfortable and engaging frame within which to author their own personal narratives: we ask them to do so not as real-world users sitting in front of a computer sending data out to a server beyond their reach, but as weary travellers telling a tale to a kindly listener.

A common theme in works of metafiction is how physical space allows a reader to traverse endless possibilities of text or time (e.g., Borges, Haig). We sought to recreate this effect in HWAMW and visually styled the interface after early text-based adventures, hypertext fiction, and visualisations of Borges' work such as Jonathan Basile's web version of the Library of Babel (libraryofbabel.info). For now, these influences are clearest in the dark theme, monospace font, garbled text animations, and the flavour of button text and other UI elements, but plans for a stronger spatial component and visual elements like ASCII art are described in Section~\ref{sec:limits-and-future}.

\subsubsection{AI outputs as ``windows'' onto existing narratives}

Central for preserving users' voices and ownership over their own stories is the way LLM-generated outputs are framed during the experience. In HWAMW, the AI-generated variations of a user's story are presented to them as windows through which they can choose to look in order to see their story differently. Metaphors for AI, like `chatbot,' `assistant,' or even `artificial intelligence' itself, reveal some qualities of the technology while obscuring others and can strongly shape a user's perception of the system \cite{khadpe2020metaphor}. Good metaphor design has the potential to suggest better applications for AI and to mitigate some of its risks \cite{blythe2025artificial}.

We find `window' to be a productive and apt metaphor for LLMs in the context of writing and, specifically, re-writing. It positions the model not as an author or originator of a text but rather as a device through which to see something differently. And it does not promise an authoritative or truthful view: much like Shannon Vallor's metaphor of the `AI mirror,' windows also have a tendency to warp or distort the light which passes through them \cite{vallor2024ai}. In HWAMW's restorying scenario, these windows each represent a particular literary influence or inspiration, surfacing rather than glossing over the derivative nature of LLM-generated text. Finally, a window is inanimate, avoiding the deceptive anthropomorphism of popular LLM-based chatbot systems.

\subsection{System Architecture} \label{architecture}

HWAMW is implemented as a next.js web application, with the main experience dynamically loaded on a single page as the user progresses through the story. Dynamic text components, like the restoried ``window'' narratives, are generated using Llama-3.3-70B-Instruct \cite{llama3}, locally hosted at University of Edinburgh. User inputs, responses, and LLM-generated stories and summaries are storied in a SQLite database for analysis.

The current prototype only supports one play-through per unique participant ID and does not allow users to return to earlier steps to edit their choices; however, we envision later versions supporting persistent user data across multiple runs as well as the ability to explore a greater number of windows in each, possibly including their own custom user-defined window interventions.

\subsubsection{Windows} \label{windows}
Each restorying intervention is implemented as a system prompt listing the title and author of an iconic work of literature in the public domain.\footnote{One work, Langston Hughes' ``Salvation,'' was mistakenly sourced from a Canadian dataset and has since been removed from the live version of the project as it is not yet public domain in the UK.} Each work is paired with a list of short ``Style rules'' and ``Structure/Content rules'' meant to recast the user's story in the style of that piece. For example, the narrative intervention for the window inspired by Jane Austen's \textit{Pride and Prejudice} is as follows:
\begin{enumerate}
    \item Style rules:
    \begin{itemize}
        \item Use witty, elevated diction with Regency flavor.
        \item Write in third person, but use free indirect discourse to flavor the narration with characters' tones and opinions.
        \item Use lots of verbal irony to tastefully critique other characters, situations, and wider society.
    \end{itemize}	
    \item Structure/content rules:
    \begin{itemize}
        \item The protagonist begins in an unfortunate or underprivileged situation that gets much worse throughout the story before being resolved.
        \item Most of the plot progresses through dialogue and changing social or familial situations.
        \item Many of the protagonist's problems are the result of misunderstandings or quick judgments; questioning their own assumptions brings growth and resolution.
    \end{itemize}
\end{enumerate}
These rules are appended to a system prompt, along with a summary of the user's story and other details they have provided throughout the interaction about themselves (e.g., motivations, traits) and other characters in their story.

For this prototype, five sample works were chosen for their popularity and distinct prose style, and the rules were developed by the authors. Full prompts and narrative intervention rules for each window used in the study are included in Appendix~\ref{app:window-defs}. See Section~\ref{sec:limits-and-future} for the limitations of this approach and future plans to diversify the interventions and ground them in relevant literary expertise.

\subsection{User Experience} 
\label{ux}

\begin{figure*}
    \centering
    \includegraphics[width=.75\textwidth]{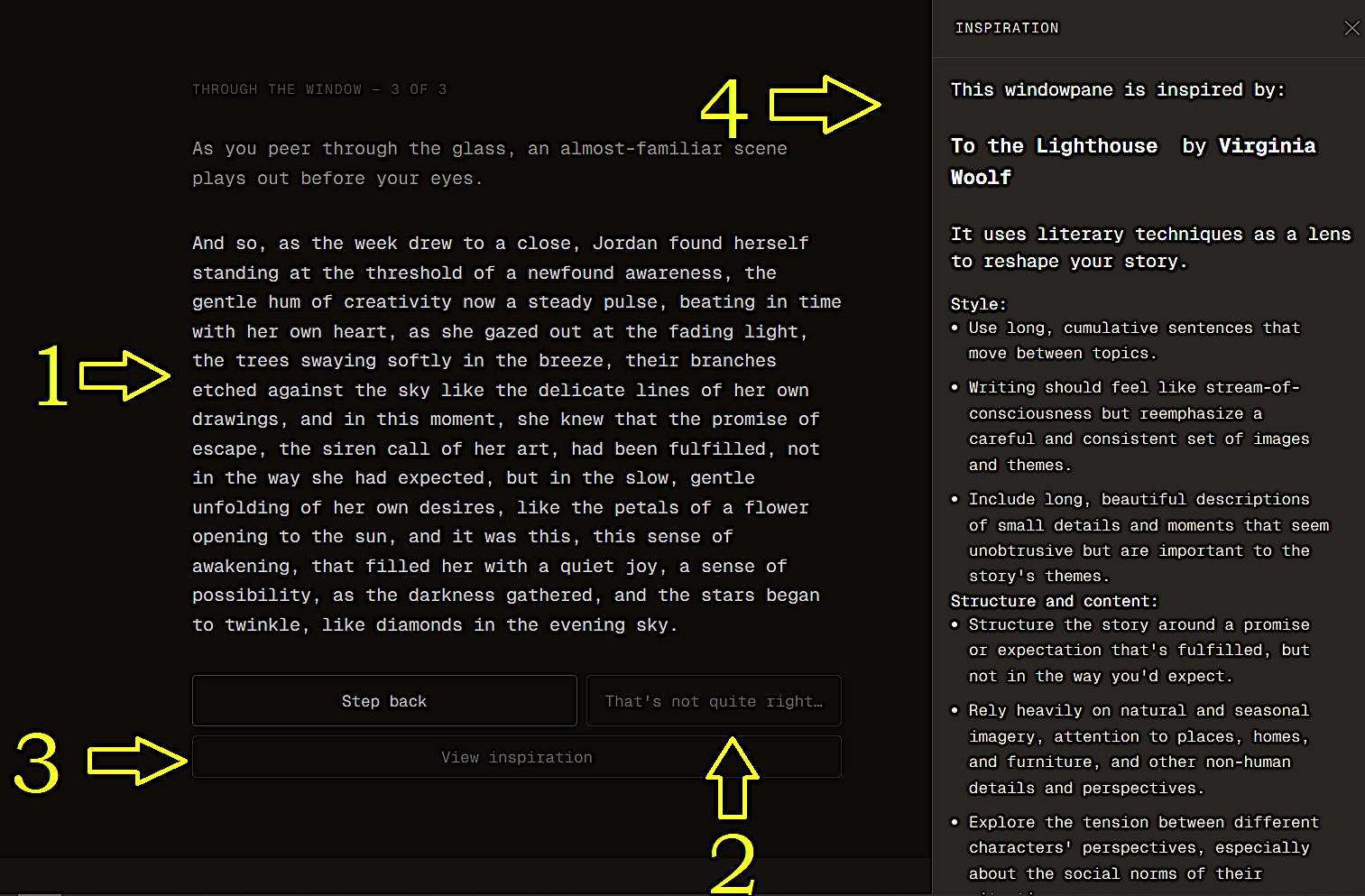}
    \caption{\textit{A screenshot from the final page of a ``window'' narrative}. The main text displays an LLM-generated variation of the user's story (1), which they can correct or amend by pressing a button and typing their requested change (2). At the end of each window narrative, users can 'View Inspiration' (3), which reveals a side panel showing the work that inspired the retelling, along with the specific literary techniques used in the system prompt to create the style or effect (4). }
    \label{fig:window}
    \Description{A screenshot of an interface with white text on a black background. The centre text is the end of a story about a user, 'Jordan', written in long, flowery sentences. Below the story are buttons the user can use to correct the story, view its inspiration, or step back. On the right, a side panel reveals that this retelling was inspired by Virginia Woolf's novel \textit{To the Lighthouse} and describes the six style and structure constraints used to retell it.}
\end{figure*}

HWAMW's interface is largely textual, inspired by electronic literature and text-based adventure games (Figure~\ref{fig:window}). The user's session unfolds as a single continuous story with three sections. We provide an example user experience and narrative here (see Appendix~\ref{app:ex-walk-through} for a full walk-through with example user inputs). 

\paragraph{The Approach}
As the story opens, our user finds themselves walking through the woods late at night, restless and grappling with their thoughts. They are presented with reflective prompts about their worries, motivations, and self-perception, framed as an inner monologue in direct address. This gives the system a range of characterising details to draw on when generating variations of their story in the second section, to reduce hallucinations and create more compelling and believable narratives. Eventually, the user discovers they are lost and soon arrive at a mysterious house. They are met by a character called the Gatekeeper, who offers them shelter in exchange for a true story (``of no more than 300 words, with a beginning, middle, and end''). The Gatekeeper presents the user with a brief LLM-generated plot summary, asking them to confirm its accuracy then following up about key characters. He asks the user to rate the story's meaningfulness on a seven-notch vertical scale (from ``A story that changed everything'' to ``A story still waiting to make sense'')---an in-story self-report framed as the Gatekeeper's guest book.

\paragraph{The House}
Then the user enters the House itself. It is filled with windows, no two precisely alike. In the prototype, they are presented with a choice of three windows (each detailed with a written description). Once the user picks a window to approach, the window's glass clouds, before clearing to reveal their own story retold in the style and structure of a well-known literary work. For example, a window described as ``a tall, sashed window richly draped with curtains,'' with laughter and quadrille music behind the glass, will present the user with a version of their story inspired by Jane Austen's \textit{Pride and Prejudice} using the interventions described in Section \ref{windows}. Although generated in full behind the scenes, the narrative is presented to the user in three shorter and more manageable parts. After reading each separate part, the user can accept the story or request a change (e.g., due to dislike or factual inaccuracy). This feedback causes the current and any future sections to be regenerated according to their suggestion. After the end of each retelling, the user can choose to view the inspiration behind the window, revealing both the work which inspired it and the explicit style and structure/content rules used to produce the transformation.

\paragraph{The Departure}
After going through all three windows, dawn arrives, and the user steps back outside. They are met again by the Gatekeeper---about to carve their name on the gatepost---when he notes that ``things always look different in the morning around here'' and asks if they would like to retell their story. The user offers another narrative (of up to 500 words). They are given access to the texts generated in each window for reference, alongside prompts emphasising that this version should be their own. The Gatekeeper then asks them to place the new story on the same seven-notch meaningfulness scale.


\section{Empirical Validation}
\label{empiricalValidation}

We conducted a study to provide empirical evidence for the psychological impacts of the HWAMW interface. While HWAMW itself was not intended to be a controlled experiment, we designed a within-subjects experiment to see we observed any statistically significant differences within individual participants, comparing before and after their experience of HWAMW.

\subsection{Recruitment} We recruited participants via the Prolific online recruitment platform (N=40; 19 female, mean age = 37.3). Participants were residents of the UK or Ireland, with at least a high school diploma, and a Prolific approval rating at or above 90\%. They were not screened for expertise in narrative or AI. All participants were presented with an AI usage declaration, in which they were asked not to use external AI tools to complete the study, and assented by typing ``I agree not to use AI tools.'' Participants were paid an average of £23.42 per hour (target pay: £20/hour; median completion time: 52 minutes) to reflect our request for them to engage in thoughtful, personal writing. All participants provided informed consent.

\subsection{Self-report measures and procedure} We asked participants to self-report their degree of familiarity with AI tools, including their frequency of use and their general feelings toward AI (e.g., excitement about potential, concern about impact on human creativity), as well as their expertise in narrative (e.g., a degree in a relevant field like literature, or deriving a portion of their income from written publication). We also asked participants to fill out the Narrative Identity Self-Evaluation (NISE) scale \cite{lind2024development}, both before and after completing the main HWAMW interface. Immediately after finishing the interface (before the second NISE scale), we surveyed participants for general feedback about their experience with the HWAMW interface. These procedure was implemented in the Qualtrics survey platform, with an external link to the HWAMW interface. We chose the NISE as our sole survey instrument because of its thematic relevance for this work; we did not use other scales, in part because we wanted to look for targeted effects with the most directly relevant instrument, rather than using a range of questionnaires in hopes that at least one yielded significant results.

\begin{figure*}
  \centering
  \includegraphics[width=.75\textwidth]{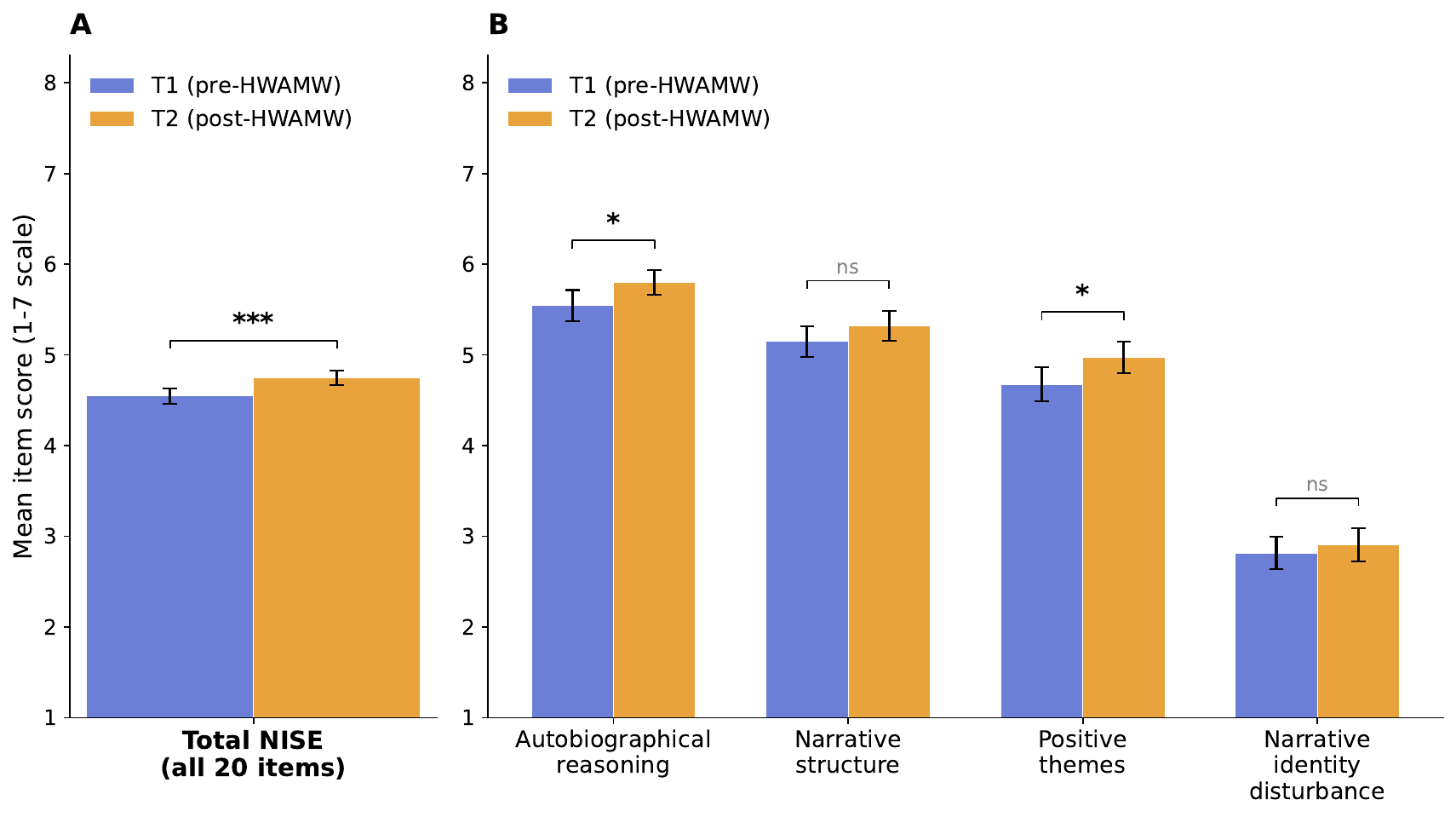}
  \caption{\textit{Narrative Identity Self-Evaluation ~\cite{lind2024development} survey scores before and after experiencing HWAMW}. (A) Participants showed a significant increase in self-reported narrative identity after using the HWAMW interface. (B) These data suggest two of the subscales seemed to drive this effect---autobiographical reasoning and positive themes.}
  \Description{Two bar plots. Plot (A) compares Total NISE scores on all 20 items, with a three star significant increase from before to after, albeit with only a visually minor increase. Plot (B) shows results of the subscale. Narrative structure and narrative identity disturbance are not significant. Autobiographical reasoning and positive themes show a one star significant increase.}
  \label{fig:user-study-results}
\end{figure*}

\subsection{Results} We analysed changes in participant responses to the NISE scale \cite{lind2024development} before and after experiencing the HWAMW interface (Fig.~\ref{fig:user-study-results}); their answers in a post-hoc user experience survey; and changes in Time 1 (pre-House) vs Time 2 (post-House) narratives, using LIWC-22 \cite{boyd2022development}. 

\paragraph{Participants showed an increase in self-reported narrative identity.} The Narrative Identity Self-Evaluation scale measures how strongly an individual relies on stories to construct their sense of self \cite{lind2024development}---based specifically on foundational research in narrative psychology \cite{mcadams2001psychology, mcadams2011narrative}. For the overall NISE scale, participant scores significantly increased from before the HWAMW interface (\textit{M} = 4.55, \textit{SD} = 0.53) to after (\textit{M} = 4.75, \textit{SD} = 0.51; \textit{t}(39) = 4.24, \textit{p} < 0.001 [Holm-corrected], \textit{d}z = 0.67; Figure~\ref{fig:user-study-results}). This increase appears to have been driven primarily by the subscales of ``autobiographical reasoning'' (\textit{t}(39) = 2.52, \textit{p} = 0.048 [Holm-corrected], \textit{d}z =0.40) and ``positive themes'' (\textit{t}(39) = 2.67, \textit{p} = 0.044 [Holm-corrected], \textit{d}z = 0.42).

\paragraph{Participants reported a generally positive experience with HWAMW} We asked participants to rate how well four statements reflected their experience with the HWAMW interface on a continuous 0--100 scale (``Overall, how well do these statements describe your experience?''). Participants rated their experience positively overall, finding it enjoyable (\textit{M} = 83.8, \textit{SD} = 17.7), thought-provoking (\textit{M} = 78.6, \textit{SD} = 29.3), and helpful for thinking about their personal stories and experiences (\textit{M} = 87.3, \textit{SD} = 17.2). Most participants also indicated they would like to use the interface again in the future (\textit{M} = 76.3, \textit{SD} = 24.1). We also asked whether they thought others would enjoy or benefit from the experience; a majority of participants (92.5\%, 37/40) indicated they could think of at least one person, group, or wider audience with whom they would consider sharing it.

\paragraph{Participants showed significant differences in their Time 1 and Time 2 stories.}

The Linguistic Inquiry and Word Count (LIWC-22) is an analysis package designed to identify psychological states as revealed in writing \cite{boyd2022development}. On the three narrative metrics in LIWC-22, participants' Time 2 narratives scored higher than their Time 1 narratives: on both Staging (expected placement of details and expository information; \(M_{T1} = 21.27\), \(M_{T2} = 32.62\)), and Plot progression (expected placement of words which signal action and character; \(M_{T1} = 9.16\), \(M_{T2} = 28.27\); Table~\ref{tab:narrativity}). There was no increase in mean Cognitive Tension scores (expected placement of words which signal internal conflict; \(M_{T1} = -4.56\), \(M_{T2} = -6.69\)). (Eight Time 1/Time 2 story pairs were excluded from this analysis as one or both stories fell below our minimum threshold of 100 words ($N_{final}=32$.)

Across the full 40 participants, we see other general trends. Sentiment analysis using VADER \cite{hutto2014vader} shows that Time 2 narratives are more positive on average. However, emotion-specific scoring with SentiArt, a sentiment analysis model trained for literary contexts \cite{jacobs2019sentiment}, shows increases in all six emotional dimensions, not just positive ones---meaning that participants' T2 narratives appear to include more affective content overall (Table \ref{tab:sentiment}).

\begin{table}[]
\caption{Sentiment Analysis of Time 1 (T1) and Time 2 (T2) narratives using VADER and SentiArt (SentiArt) lexicons.}
\label{tab:sentiment}
\resizebox{\columnwidth}{!}{
\begin{tabular}{ll|l}
                              & \textbf{T1} & \textbf{T2}                                                                     \\
Sentiment (VADER)                   & \begin{tabular}[c]{@{}l@{}}mean: 0.3154  \\ std: 0.7502 \end{tabular}  & \begin{tabular}[c]{@{}l@{}} mean: 0.6101  \\ std: 0.6262 \end{tabular} \\ \hline
Affective-Aesthetic Potential (SentiArt)  & \begin{tabular}[c]{@{}l@{}}mean: 0.0565  \\ std: 0.1624 \end{tabular}  & \begin{tabular}[c]{@{}l@{}} mean: 0.1360  \\ std: 0.2012 \end{tabular} \\ \hline
Happiness (SentiArt)  & \begin{tabular}[c]{@{}l@{}}mean: 0.8112  \\ std: 0.2068 \end{tabular}  & \begin{tabular}[c]{@{}l@{}} mean: 0.9034  \\ std: 0.2329 \end{tabular} \\ \hline
Sadness (SentiArt)  & \begin{tabular}[c]{@{}l@{}}mean: 0.6519  \\ std: 0.1375 \end{tabular}  & \begin{tabular}[c]{@{}l@{}} mean: 0.7906  \\ std: 0.1920 \end{tabular} \\ \hline
Fear (SentiArt)  & \begin{tabular}[c]{@{}l@{}}mean: 0.8191 \\ std: 0.2033 \end{tabular}  & \begin{tabular}[c]{@{}l@{}} mean: 0.8650  \\ std: 0.2222 \end{tabular} \\ \hline
Surprise (SentiArt)  & \begin{tabular}[c]{@{}l@{}}mean: 0.4684 \\ std: 0.1527 \end{tabular}  & \begin{tabular}[c]{@{}l@{}} mean: 0.5423  \\ std: 0.1663 \end{tabular} \\ \hline
Anger (SentiArt)  & \begin{tabular}[c]{@{}l@{}}mean: 0.3564 \\ std: 0.1502 \end{tabular}  & \begin{tabular}[c]{@{}l@{}} mean: 0.4723  \\ std: 0.2022 \end{tabular} \\ \hline
Disgust (SentiArt)  & \begin{tabular}[c]{@{}l@{}}mean: 0.2259 \\ std: 0.1100 \end{tabular}  & \begin{tabular}[c]{@{}l@{}} mean: 0.2751  \\ std: 0.1473 \end{tabular} \\ \hline
\end{tabular}
}
\end{table}

\begin{table}[]
\caption{LIWC-22 Narrative Metrics (N=32). For narrativity scores, 100 represents a typical curve; -100, its inverse; 0, an atypical curve}
\label{tab:narrativity}
\resizebox{\columnwidth}{!}{
\begin{tabular}{ll|l}
                              & \textbf{T1}  & \textbf{T2}                                                                      \\
Wordcount                     & \begin{tabular}[c]{@{}l@{}}mean: 216.15625 \\ std: 69.918897\end{tabular}  & \begin{tabular}[c]{@{}l@{}}mean: 242.71875  \\ std: 81.916651\end{tabular} \\ \hline
Narrativity\_Overall          & \begin{tabular}[c]{@{}l@{}}mean: 8.620000\\ std: 46.795707\end{tabular}    & \begin{tabular}[c]{@{}l@{}}mean: 18.065938\\ std: 48.718731\end{tabular}   \\ \hline
Narrativity\_Staging          & \begin{tabular}[c]{@{}l@{}}mean: 21.270625  \\ std: 61.374223\end{tabular} & \begin{tabular}[c]{@{}l@{}}mean: 32.619687\\ std: 65.242111\end{tabular}   \\ \hline
Narrativity\_PlotProgression  & \begin{tabular}[c]{@{}l@{}}mean: 9.1575  \\ std: 60.438172\end{tabular}    & \begin{tabular}[c]{@{}l@{}}mean: 28.2650  \\ std: 63.941620\end{tabular}   \\ \hline
Narrativity\_CognitiveTension & \begin{tabular}[c]{@{}l@{}}mean: -4.564688  \\ std: 71.033239\end{tabular} & \begin{tabular}[c]{@{}l@{}}mean: -6.686562  \\ std: 67.881641\end{tabular}
\end{tabular}
}
\end{table}

\section{Expert Review}
\label{expertReview}

With this empirical validation of HWAMW's engagment and potential effects on user's narratives as well as their narrative identities, we sought to better understand how users experienced this effect and how it might be improved in future iterations of HWAMW's design. To do so, we conducted an expert review study to elicit and qualitatively analyse in-depth responses from researchers with relevant expertise. The HWAMW interfaces draws on a range of relevant disciplines---from HCI and narrative to AI and behavioural science. Accordingly, we sought feedback from experts with backgrounds in both AI and narrative studies who might see different approaches and opportunities for the system.

\subsection{Recruitment} We recruited participants via our personal network of collaborators and academic peers (N=10; 8 female). Participants were selected on the basis of having relevant expertise in both AI and narrative (or adjacent fields) at a postgraduate or equivalent level. Participants were paid £30 for their participation. All participants provided informed consent.

\subsection{Procedure} Unlike the general validation study, we did not ask our expert users to respond to any self-report measures prior to the HWAMW interface. Instead, they were presented with a structured free-response survey posing open-ended questions about their experience, the framing and purpose of the system, the LLM-generated stories, and the overall design approach. Responses were coded using template analysis \cite{king200421} to develop both targeted and emerging themes and to capture feedback and suggestions to inform future work.

\subsection{Results} Expert user responses, and the themes we developed around them, encompass a broad range of considerations and suggestions. Here, we present the results most relevant to our research questions. All quotes in this section are copied directly from user responses, apart from marked edits for contextualisation and abbreviation.

\subsubsection{Restorying with HWAMW challenged and expanded users' stories while retaining their voice.} The experts described having experiences characteristic of restorying. Writing, then rewriting, their stories helped them to work through blocks they encountered when writing about particular experiences, ``think through the incident with more detail,'' and ``consider something more'' than what had occurred to them in the initial telling. Beyond just rewriting, they also valued the iterative and multiple nature of the restoried ``windows'' they encountered along the way, productively distancing themselves from their own stories, ``disrupting habitual ways of seeing a story and opening up directions that might otherwise remain invisible.'' The windows themselves were frequently discussed as enabling them to see their story from other perspectives, and many described seeing the meaning of their story---its continued relevance, its worth, or what it was ``about''---shift as a result. Crucially, they still felt agency and ownership over their stories: one described the windows as helping them to ``find [their] own voice perhaps in opposition to famous others,'' and another described the final story and perspective as still feeling ``very my own.''

\subsubsection{HWAMW's narrative framing made it easier to engage.} Expert users found the metafictional setup highly engaging and imaginative. One commented specifically that they enjoyed ``the justification for why I'm doing the exercise as opposed to using the tool just because;'' another described it as ``[allowing] perspective change to be experience narratively rather than presented as explicit advice or interpretation.'' They saw HWAMW as distinct from journaling systems, creative writing tools and exercises, and video games, though related to each: the narrative framing created a space for guided engagement without the explicit guidance of therapy. The narrative frame also made engagements involving personal or vulnerable stories feel more natural: one expert stated that ``the humorous framing going into the `world' created was extremely valuable'' and made them ``more comfortable sharing personal information and buying into the whole narrative experience.''

\subsubsection{Window interventions based on iconic works are engaging, but could be expanded.} The experts generally enjoyed reading variations of their stories inspired by iconic works of fiction. In addition to the novelty of some styles and the fun they had encountering works they recognised, a few mentioned that this approach connected them to wider cultural meanings and resources---``a kind of literary genealogy,'' as one described it, or inspiration and implicit additional resources (i.e., motivation to read the source texts) to another. However, they each suggested some broadening of the range of interventions, from including works from a wider range of genres, forms, and writers (especially those which challenge limiting notions of literary canon) to considering a more radical set of interventions, including dramatic changes in emotional tone, temporality, plot, and setting. At the same time, many acknowledged that even the current interventions may not be suitable for all users' stories (in particular, an Edgar Allen Poe-inspired window which tended toward melancholy and despair) and suggested that certain windows might come with warnings or have the nature of their intervention indicated to the user in some way.

\subsubsection{LLM limitations were obvious to every user, but surprisingly generative for some.} Finally, the generated texts struck several users as predictable and unsurprising. Most commented on the cliché or overwrought nature of many of the windows, particularly those emulating writers with a richly descriptive prose style. For some, this was obtrusive, annoying, or even offensive given the nature of their story, although some also found the overblown prose entertaining. Counterfactuals and hallucinations were even more divisive: several users preferred realistic windows which faithfully relayed events in a tone which felt natural. Others found moments of mismatch between generated narratives and their own experiences valuable---in some cases, funny; in others, reflectively useful, ``as it could provoke readers to disagree, and in doing so, think more about what the truth actually is.'' For one user, this was a defining pattern of the experience: ``plausibility did not always determine usefulness;'' and the tonal mismatches ``unexpectedly helped me locate my own emotional position more precisely.'' 

Regardless of the model's shortcomings, experts saw HWAMW as a promising use for LLMs. One, who described themself as a ``generally anti-AI person,'' found that their view of generative AI in creative writing was expanded by the nuanced way HWAMW used generated stories not their own to help them produce one which was. Another was interested in ``the shift from AI as a producer of narrative to AI as an instrument for perceiving narrative. The generated text does not have to be the final creative product; its value can lie in making structures, assumptions, or possibilities in the human story newly visible.''

\section{Discussion}
\label{discussion}

A major theme of AI discourse is the concern that AI-assisted writing often takes away from human writing more than it adds to it. Thrown into society as a general purpose tool without explicit design, AI-assisted writing has the potential to erode much of what we find valuable in storytelling---both in terms of the stories we create and the process of generating them. We present The House with a Million Windows (HWAMW) interface as one potential means of designing AI-assisted writing systems in a way that layers meaning and intentionality onto the telling of our stories, rather than stripping them away.

Empirical evidence suggests that, even as a prototype, HWAMW helps users with and without narrative expertise to re-encounter and reinterpret their stories (Section~\ref{empiricalValidation}). We suggest that its metafictional framing, interactive elements, and careful framing of both human- and AI-generated texts fill a valuable and under-explored space in the design of writing technologies \cite{lee2024design} where users can explore their own work in the freedom of a fictional setting, using voices which are \textit{not} their own to refine and develop one which authentically \textit{is}. Our empirical validation suggests that they may find more (and different) meaning in their stories in the process.

 While prior work focused on the benefits of a particular narrative intervention, such as the Hero's Journey \cite{rogers2023seeing}, HWAMW invites a broad range of literary styles and techniques which may be valuable within this process and makes these accessible even to users without specific narrative training or experience. The Hero's Journey may be a powerful megamyth, but it is still a limited structure which may not suit every story---it assumes a happy ending, for instance, and imposes a certain level of individualism by centring the protagonist as the hero of their tale. HWAMW's ``windows,'' while currently numbering far less than a million, suggest a larger space of possible stories which might connect users not only to new perspectives but to wholly new layers of cultural meaning as they reinterpret their tales according to different styles, chronologies, and artistic movements. This advances the vision for Interpretive Technologies as systems designed around interpretive depth and cultural complexity \cite{kommers2026computational,hemment2025doing}. While our current experimental setup does not allow for direct comparison with previous psychological findings, we do find promising indications that our wider approach to restorying, based in literary traditions of homage and adaptation, can also significantly increase users' sense of narrative identity.

Beyond helping users retell their stories, HWAMW also demonstrates the potential to productively restory the role of AI technologies in human storytelling. Generative AI is an inherently cultural technology, both arising out of and contributing to the contexts in which it is developed \cite{kommers2026computational}. This hermeneutic approach extends to the stories we tell with, about, or in relation to LLMs---in some sense, they are born out of our stories. Mark Coeckelbergh writes that understanding AI technologies as both products and co-producers of narratives ``removes humans from their supposedly central position as sole narrators of their stories (technology also narrates)'' and acts as ``a stimulus to both critical analysis and change [...] We can co-create new, better processes and stories'' \cite{coeckelbergh2021time}. 

We suggest that interfaces like HWAMW have considerable potential to give humans more agency over both kinds of story: those told about us, and those told about AI. As our expert users suggested, HWAMW's design facilitates the creation of better stories \textit{for} AI: uses which position it not as a tool to automate or replace human storytelling but as a means for encountering, challenging, and re-encountering our narratives. By positioning LLMs as counter-narratives we can use to better understand our own, interfaces like HWAMW can acknowledge generative AI's cultural and contingent nature and ultimately use it to help humans find greater meaning in the stories we tell.

\section{Limitations and Future Work}
\label{sec:limits-and-future}

\paragraph{Breadth and depth of windows: The prototype includes only a small number of windows, drawn from a limited notion of literary canon.} Future work can include an expanded number of literary interventions. Based on expert feedback, future work can incorporate a more diverse range of authors and works as well as a broader set of interventions, including windows based in additional subgenres (e.g., science fiction) and forms (e.g., prose poetry) and windows implementing a specific narrative technique, like a shift in point-of-view or a non-linear plot. Furthermore, future preset windows can be co-designed with domain experts (e.g., scholars or practitioners of a specific genre or style), and users will be supported in designing and testing window interventions of their own. Future work can also investigate the effects of individual windows---whereas in this prototype we prioritised an integrated user experience rather than experimental precision in measurements of effects from specific windows. 

\paragraph{Breadth and depth of psychological benefits: The validity looks at only one psychological scale} We find that participants showed an an increase in ``narrative identity'' after using HWAMW, based on a self-report scale. This is useful, as it provides evidence that some relevant psychological effects are elicited in a way that is consistent with the intentions of the design. But further research is required to interpret these findings and determine what they mean in practice. We plan to more closely examine the psychological effects of HWAMW by exploring other psychological scales, conducting targeted window-by-window comparisons, and incorporating richer qualitative questions and closer readings of restoried texts into future analyses. In particular, we are interested in exploring to what extent HWAMW's effects are linked to its overall design and metafictional framing, its restorying format, and the particular windows a user might encounter. We also consider the durability of any psychological effects: under what conditions might they exert a lasting impact on an individual's perception of meaning, rather than being directly tied to the momentary use of the HWAMW system?

\paragraph{Visual/auditory design: The current prototype is solely text-based} The House with a Million Windows offers a metaphor with an implication of rich imagery. Future iterations will explore more immersive visual design and spatial navigation. We also see potential to explore multimodal inputs and outputs, such as spoken stories. This could make the system more accessible to users for whom longform writing is difficult or intimidating and may also change the kinds of stories they tell and the nature of their restorying experience.

\paragraph{Specific applications: The current prototype only looks at intrapersonal meaning} In this prototype, we target a psychological notion of meaningful experience, but HWAMW could be deployed for a range of other target outcomes. For example, windows might encourage users to consider variations in cultural meaning, from the perspectives of people with different backgrounds or worldviews, rather than centring their individual psychological experience. Stories might also be defined and reworked in terms of their real world impacts: for example, restorying narratives concerning climate change, technology, or misinformation.

\begin{acks}
This work was supported by the Alan Turing Institute under Lloyd’s Register Foundation grant ATI/100004. 
\end{acks}

\section*{Ethics and Privacy Statement}
This study was approved by Edinburgh Informatics Research Ethics Process, reference number 921742. In addition to the risks considered by the review board, we acknowledge that the interface presented in this paper bears resemblance to certain forms of therapy and risks misinterpretation and misuse. HWAMW is a narrative experience, not a form of therapy, and it does not have any clinical aims or oversight. As with all LLM-based technologies, its outputs are unpredictable and subject to a degree of randomness. It should not be used by psychologically vulnerable populations or treated as a therapy tool. We have taken care to communicate this in the design of the system and our reporting of it here. While we plan to make our code publicly available, we ask that anyone choosing to deploy the application or build on its design warns users of the risks of using LLMs and takes measures to safeguard their deployment against misuse.

\bibliographystyle{ACM-Reference-Format}
\bibliography{references}


\appendix

\section{Example User Walk-through} 
\label{app:ex-walk-through}

We provide a walkthrough of the narrative from a user's perspective, using pseudonymised (and occasionally abbreviated) real inputs from one of our users, who we'll refer to as Ahmed.

\subsection{The Approach}
HWAMW opens with the user walking in the woods late at night, restless and grappling with their thoughts. Their character's physical journey prompts them to reflect on their worries: ``You're exhausted, but you keep walking, as though your feet moving on might make your thoughts follow after.'' The bottom of this page asks them to list their woes from a multiple-selection list (including an ``other''/opt-out selection labelled ``You wouldn't understand''). Ahmed selects ``My career,'' ``My creative pursuits,'' and ``My community.''

As the walk continues, the text describes a growing sense that he is wandering somewhere he should not be, while the reflective prompts continue to learn more about his situation. In three free-text description boxes, he enters that he is ``Caring,'' ``Affectionate,'' and ``Cheerful,'' then follows up with more detail: ``[...] I give my all to my family and my loved ones.'' These details will provide the system with helpful context to draw on in later steps, but, in the moment, they serve as reflective prompts and help Ahmed think about the story he would like to tell.

Eventually, the narrative leads Ahmed to the realisation that he is lost, with nowhere to turn except to a mysterious house in the distance. A kindly Gatekeeper listens to his plight and invites him to stay the night, as long as he is willing to tell a true story first. The Gatekeeper instructs that it should be something true, with a beginning, middle, and end, and no more than 300 words. Ahmed responds with three short paragraphs about his career in sports. He had always been interested in football, but discovered a new passion during university: ``This perspective initially came through playing a different type of video game call [sic] FIFA Manager. I discovered a new world of football, the intricacies behind it and i instantly fell in love with it.'' He describes managing his first team and loving the relational and personal aspects of the job. ``Issues however crept up along the line with threatened the harmony of the team [...] I took the the decision i felt was best for the team [...] and stepped aside for my assistant to lead the charge. The team went on to win the competition, the first and only during our undergraduate days.''

When Ahmed submits this story, a brief loading screen explains that the Gatekeeper is considering thoughtfully. The Gatekeeper then presents him with an LLM-generated summary of the basic plot of his story. Ahmed confirms that the details are correct, then supplies some more detail about his assistant manager as the Gatekeeper follows up about the other characters in his story. (The system prompts for details for up to three important characters, besides the user, but many users, like Ahmed, had few other named characters in their story.) Finally, the Gatekeeper asks what his motivations are---``What are you hoping to get out of this, exactly?'' Ahmed selects several options, including ``To lead a stable, happy life,'' ``To support my family, friends, and community,'' and ``To make a difference in the world.'' Given the option to elaborate, he answers ``That's basically it. I am a simple person.''

The Gatekeeper then asks Ahmed how meaningful his story is, explaining that he keeps a guest book, of sorts, on the gate post, and he carves the names of guests with the most meaningful stories at the top. Ahmed is presented with a vertical input with seven notches, ranging from the most meaningful at the top (``A story that changed everything'') to the least meaningful at the bottom (``A story still waiting to make sense''). He selects the third slot, just one above the center. The Gatekeeper also asks him for his name and preferred pronouns, then welcomes him into the House.

\subsection{The House}
A longer text passage describes the odd, gallery-like space Ahmed encounters once inside. The whole building is full of windows, no two alike, shining with odd light that does not seem to come from the night sky outside. After reading through a description of his initial wander through the building, Ahmed continues on to approach some of the windows nearest him.

Here, the interface presents him with three options, each a vivid description of the window itself. He selects ``A tall, sashed window richly draped with curtains. You can hear coy laughter and the faint notes of a quadrille coming through the glass.'' The interface loads for a few seconds, describing the glass of the window beginning to cloud, then clear, before displaying the first section of ``an almost-familiar scene'' playing out before his eyes:
\begin{quote}
    It was a truth universally acknowledged, that a young man in possession of a fervent passion for football, must be in want of a means to express it. Ahmed, a university student of modest means and unassuming demeanor, had once found solace in the beautiful game, his feet moving in tandem with the ball as if by instinct. However, as he delved into the realm of FIFA Manager, his perspective underwent a profound shift[...]
\end{quote}
After two more of these 100-150-word sections, the story in this window concludes, and Ahmed can click a button to view the inspiration behind the retelling. In this case, he learns that the window is based on \textit{Pride and Prejudice} by Jane Austen, and that the effect was achieved with the following rules:
\begin{itemize}
    \item Style rules:
    \begin{enumerate}
        \item Use witty, elevated diction with Regency flavor.
        \item Write in third person, but use free indirect discourse to flavor the narration with characters' tones and opinions.
        \item Use lots of verbal irony to tastefully critique other characters, situations, and wider society.
    \end{enumerate}	
    \item Structure/content rules:
    \begin{enumerate}
        \item The protagonist begins in an unfortunate or underprivileged situation that gets much worse throughout the story before being resolved.
        \item Most of the plot progresses through dialogue and changing social or familial situations.
        \item Many of the protagonist's problems are the result of misunderstandings or quick judgments; questioning their own assumptions brings growth and resolution.
    \end{enumerate}
\end{itemize}
After stepping back from the window, Ahmed can choose two others to explore---each time selecting from a set of three he has not yet seen. Then dawn arrives, and he steps back outside.

\subsection{The Departure}
Upon leaving the house, Ahmed encounters the Gatekeeper again, this time preparing to carve his name on the gatepost. He stops on seeing Ahmed's face and asks whether the story has changed, suggesting that ``things always look different in the morning around here.'' Ahmed now has the opportunity to retell his tale, in up to 500 words this time. A button labeled `Recall what the House showed you' allows him to revisit the window texts, but the prompts above the textbox emphasise that this new story should be his own.

Ahmed's new story is slightly longer and much more descriptive: ``I recall my passion for playing football. How much i loved the game as a striker, breaking down defenses, barging down the goal post [...] I soon developed a love for a different aspect of the game. One that was still played on the pitch but this time without me lacing my boots and terrifying defenders and goalkeepers.'' He includes a clearer shift into conflict: ``This journey started off great [...] Along the line, however, individuals with egos who couldn't see beyond their noses [...] sought to put the team's harmony in jeopardy. I was not going to let all the hard work and efforts of the larger team go to naught [...]'' The ending also provides more perspective; not only does the team win the competition, but Ahmed also ended up in ``a much bigger and more important role'' within the sport.

Finally, the Gatekeeper asks Ahmed where this new version of the story belongs, and he is presented with the same vertical scale. This time, he chooses to place it at the very top.

\section{Window Definitions} 
\label{app:window-defs}
\subsection{``Hills Like White Elephants'' by Ernest Hemingway}
\textbf{Description:} The simple, unvarnished pine frame. It smells of cat hair, peat, and cigarettes.

\textbf{Style rules:}
\begin{itemize}
     \item Use short sentences.
     \item Speak plainly.
     \item Use lots of dialogue.
     \item Don't use dialogue tags (like \"he said\") to show who's speaking if the context makes it obvious.
\end{itemize}
\textbf{Structure/content rules:}
\begin{itemize}
     \item Reveal the tension subtly and only indirectly through a dialogue between two major characters.
     \item A physical journey symbolises the protagonist's difficult decision.
     \item The protagonist's mind should be made up by the story's end, but the story only implies what that decision is indirectly.
\end{itemize}

\subsection{\textit{To the Lighthouse} by Virginia Woolf}
\textbf{Description:} A comfortable frame weathered slightly by sea air. A woman's green cashmere shawl is draped over its corners, but you can see light shimmering on the glass beyond.

\textbf{Style rules:}
\begin{itemize}
     \item Use long, cumulative sentences that move between topics.
     \item Writing should feel like stream-of-consciousness but reemphasise a careful and consistent set of images and themes.
     \item Include long, beautiful descriptions of small details and moments that seem unobtrusive but are important to the story's themes.
\end{itemize}
\textbf{Structure/content rules:}
\begin{itemize}
     \item Structure the story around a promise or expectation that's fulfilled, but not in the way you'd expect.
     \item Rely heavily on natural and seasonal imagery, attention to places, homes, and furniture, and other non-human details and perspectives.
     \item Explore the tension between different characters' perspectives, especially about the social norms of their situation.
\end{itemize}

\subsection{``The Tell-Tale Heart'' by Edgar Allen Poe}
\textbf{Description: }The slender frame of burnished mahogany, gleaming bloodred in the candlelight and seeming to cast a shadow longer than it ought.

\textbf{Style rules:}
\begin{itemize}
     \item Use a strong first-person narrative voice, with lots of interiority.
     \item Use poetic, formal, and even archaic language.
     \item The tone should be dark, dwelling on Gothic imagery and themes.
     \item Use an unreliable narrator who feels logical, justified, and clinical on the surface but spirals, growing more emotional or uncertain throughout the piece.
\end{itemize}
\textbf{Structure/content rules:}
\begin{itemize}
     \item The protagonist, looking back, narrates their story in retrospect, explaining or justifying their decisions.
     \item A strong fixation--whether a sound, person, image, or idea--motivates the protagonist and underscores the main conflict.
     \item Build tension up to a dramatic climax, where a decisive action or exclamation concludes the tale.
\end{itemize}

\subsection{\textit{Pride and Prejudice} by Jane Austen}
\textbf{Description:} A tall, sashed window richly draped with curtains. You can hear coy laughter and the faint notes of a quadrille coming through the glass.

\textbf{Style rules:}
\begin{itemize}
     \item Use witty, elevated diction with Regency flavor.
     \item Write in third person, but use free indirect discourse to flavor the narration with characters' tones and opinions.
     \item Use lots of verbal irony to tastefully critique other characters, situations, and wider society.
\end{itemize}

\textbf{Structure/content rules:}
\begin{itemize}
     \item The protagonist begins in an unfortunate or underprivileged situation that gets much worse throughout the story before being resolved.
     \item Most of the plot progresses through dialogue and changing social or familial situations.
     \item Many of the protagonist's problems are the result of misunderstandings or quick judgments; questioning their own assumptions brings growth and resolution.
\end{itemize}

\subsection{``Salvation'' by Langston Hughes}
\textit{Note: will not be used in future iterations, to ensure that all works are public domain and accessible to participants in the countries where the system is deployed}

\textbf{Description:} A weary frame set with sagging hinges, giving off a soulful air. Cigarette smoke and faint hints of jazz music seep through the edges.

\textbf{Style rules:} 
\begin{itemize}
     \item Keep a conversational tone, with lots of verbal irony
     \item Draw on imagery associated with belief and disillusionment, like light and music. Emphasize moments of hopefulness, guilt, compassion, and sorrow.
     \item Use polyphony: incorporate the voices of minor characters as strong and distinct personas whose comments reflect their social role and background.
\end{itemize}
\textbf{Structure/content rules:}
\begin{itemize}
     \item Emphasize social and cultural commentary.
     \item Present the story in retrospect: a first-person narrator looking back and reflecting on events in past tense, but with close moment-by-moment attention to detail.
     \item The narrator should directly show the reader the internal conflicts that other characters can't see.
\end{itemize}

\end{document}